\documentclass[10pt,twocolumn,letterpaper]{article}

\usepackage{cvpr}              

\newcommand{\red}[1]{{\color{red}#1}}

\definecolor{cvprblue}{rgb}{0.21,0.49,0.74}
\usepackage[pagebackref,breaklinks,colorlinks,allcolors=cvprblue]{hyperref}
\usepackage{multirow}
\usepackage{tabularx} 
\usepackage{graphicx}
\usepackage[capitalize]{cleveref}
\usepackage{enumitem}
\usepackage{marvosym}

\def\paperID{3170} 
\def\confName{CVPR}
\def\confYear{2025}

\def\name{FreeSim }

\title{FreeSim: Toward Free-viewpoint Camera Simulation in Driving Scenes}

\author{
  Lue Fan$^{1,2*}$, \;
  Hao Zhang$^{2*}$, \;
  Qitai Wang$^{1}$, \;
  Hongsheng Li$^{2}$\textsuperscript{\Letter}, \;
  Zhaoxiang Zhang$^{1}$\textsuperscript{\Letter} 
  \\ 
$^1$ CASIA \; $^2$ MMLab, CUHK \\
\tt\small{\{lue.fan, wangqitai2020, zhaoxiang.zhang\}@ia.ac.cn}  \;
 \tt\small{\{1155231340@link, hsli@ee\}.cuhk.edu.hk} \\
  \small{Project page: \url{https://drive-sim.github.io/freesim}}
}

\begin{document}
\twocolumn[{%
\vspace{-2em}
\maketitle%
{
    \centering
    \includegraphics[width=0.99\linewidth]{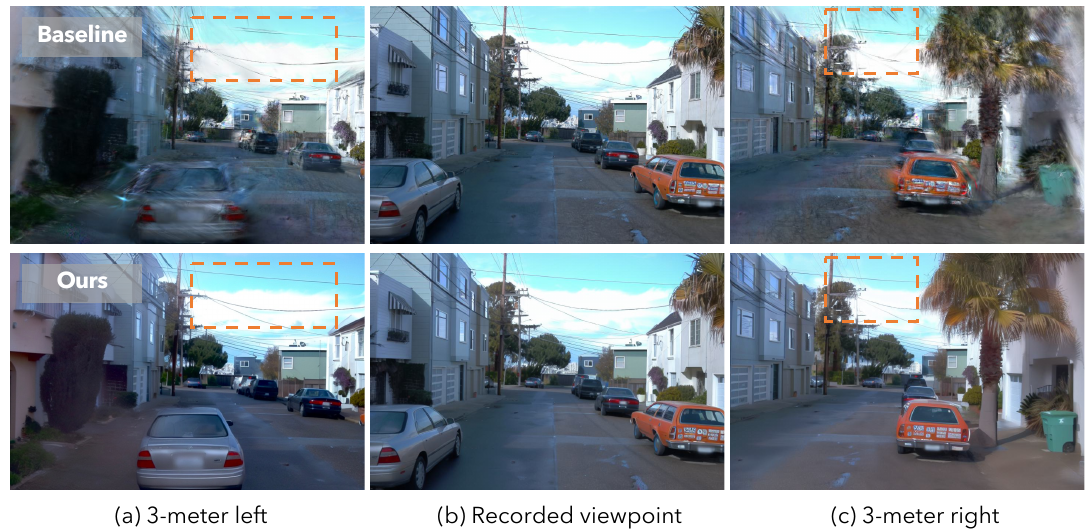}
    \captionof{figure}{The proposed \name can obtain high-quality camera simulation results in viewpoints largely deviated from the recorded trajectories. Here we adopt pioneering work PVG~\cite{chen2023periodic} as the baseline for demonstration. We highlight the delicate \emph{power lines} with \textcolor{orange}{orange} bounding boxes.}
    \vspace{3mm}
    \label{fig:teaser}
    }
}]
{\let\thefootnote\relax\footnote{$^*$Equal contribution. {\Letter} Corresponding authors.}}

\vspace{-3mm}
\begin{abstract}
We propose FreeSim, a camera simulation method for autonomous driving. FreeSim emphasizes high-quality rendering from viewpoints beyond the recorded ego trajectories.
In such viewpoints, previous methods have unacceptable degradation because the training data of these viewpoints is unavailable.
To address such data scarcity, we first propose a generative enhancement model with a matched data construction strategy.
The resulting model can generate high-quality images in a viewpoint slightly deviated from the recorded trajectories,  conditioned on the degraded rendering of this viewpoint.
We then propose a progressive reconstruction strategy, which progressively adds generated images of unrecorded views into the reconstruction process, starting from slightly off-trajectory viewpoints and moving progressively farther away. 
With this progressive generation-reconstruction pipeline, FreeSim supports high-quality off-trajectory view synthesis under large deviations of more than 3 meters.
\end{abstract}

\vspace{-5mm}
\section{Introduction}
\label{sec:intro}
Realistic simulation is widely recognized as a foundational component of embodied intelligence and autonomous driving. With the advent of 3D Gaussian Splatting (3DGS) ~\cite{kerbl3Dgaussians}, camera simulation for driving has rapidly advanced. A line of works ~\cite{luiten2023dynamic,wu20244d,yang2024deformable,huang2024s3gaussian,chen2023periodic,zhou2024hugs,chen2024omnire,yan2024street} utilizing 3DGS have achieved efficient reconstruction and high-quality rendering in driving scenes.
However, most of them only focus on reconstructing and rendering along recorded trajectories, limiting their generalizability to unrecorded \emph{off-trajectory} viewpoints (e.g., the viewpoints shifted 3 meters right in \cref{fig:teaser}).
A practical simulator, however, must generalize beyond recorded trajectories, to support high-fidelity and consistent rendering in trajectories that a self-driving car may take when executing various action decisions. 

Maintaining realism and consistency in off-trajectory viewpoints is challenging, with the core difficulty being the lack of ground-truth data for those viewpoints.
Recently, some methods  ~\cite{gao2024cat3d,liu2024reconx,zhou2023sparsefusion,voleti2025sv3d,liu20243dgs} have leveraged generative models to synthesize novel views from sparse, multi-view inputs.
However, these methods cannot directly meet our goals to generate data outside recorded trajectories, as they rely on a substantial number of multi-view training samples.
Such samples are unavailable in autonomous driving datasets, where data is confined to a single moving-forward trajectory.

To tackle this problem, we propose a generation-reconstruction hybrid method along with a data construction strategy to support the training of generative models.
To illustrate our motivation, let us assume we have a radiance field reconstructed from recorded views. If the viewpoint deviation is small, it may yield slightly degraded renderings.
It is much easier to restore a high-quality image from its slightly degraded counterpart than from a pose transformation condition like ``laterally shift the front camera for 3 meters''.
In this way, we reformulate the novel view generation task as a much easier generative image enhancement task. We then face two challenges: (1) how to train the generative image enhancement model for off-trajectory views without ground truth data, and (2) how to extend such small viewpoint deviations to larger viewpoint transformation.

Since we still lack off-trajectory ground truth, we turn to degrading on-trajectory images to simulate the rendering pattern of off-trajectory views, thus creating training pairs for the generation model.
Specifically, we first propose an efficient \emph{piece-wise Gaussian reconstruction} strategy, which partitions each full trajectory into small sub-segments.
A number of small-scale piece-wise Gaussian fields can be obtained from these sub-segments.
We then conduct \emph{extrapolated rendering} in held-out unseen future frames of each sub-segment.
The rendering of such extrapolated views can simulate the degraded rendering patterns similar to off-trajectory renderings.
We further add specifically designed noise to Gaussian primitives of these piece-wise Gaussian fields during rendering, improving the data diversity and scale.
With these strategies, we construct a large-scale dataset of 1.5 million samples to support the training of the enhancement model.

After addressing small viewpoint deviations, we extend this capability to larger viewpoint change.
To this end, we design a progressive, generation-reconstruction alternate approach that progressively adds off-trajectory views during reconstruction, from small viewpoint deviations to large viewpoint deviations.
In this process, the generative model produces high-quality images conditioned on the on-the-fly renderings of the newly added off-trajectory viewpoints.
The generated images are adopted to update the training image set.
This design avoids direct generation from severely degraded images caused by large viewpoint changes, ensuring that the reconstruction process can smoothly expand from on-trajectory viewpoints to far away off-trajectory viewpoints. We summarize our contributions as follows.
\begin{enumerate}
    \item To address data scarcity in free-viewpoint camera simulation, we reformulate the challenging pose-conditioned view generation task into a generative image enhancement task and propose a matched data construction pipeline for model training.
    \item We propose a progressive reconstruction strategy to seamlessly combine the generation and reconstruction parts, avoiding severe image degradation under large viewpoint changes. 
    \item \name achieves significant superiority to existing reconstruction-based methods in free-viewpoint rendering, moving one step closer to a realistic and practical simulator.
\end{enumerate}
\section{Related Work}

\paragraph{Reconstruction-based camera simulation in driving scenes.} 
Neural Radiance Fields (NeRF) ~\cite{mildenhall2021nerf} and 3D Gaussian Splatting (3DGS) ~\cite{kerbl3Dgaussians} are the most popular reconstruction methods.
3DGS utilizes explicit representation of 3D Gaussian and the rasterization-based rendering method, resulting in faster rendering and training.
Both methods can achieve high-quality, photorealistic novel view synthesis (NVS) given a few images of some 3D scenes taken from different camera viewpoints. The later works ~\cite{yang2023emernerf,liu2023real,tonderski2024neurad,yang2023unisim,wu2023mars,tonderski2024neurad} for NeRF and ~\cite{zhou2024drivinggaussian,wu20244d,yang2024deformable,huang2024s3gaussian,chen2023periodic,zhou2024hugs,chen2024omnire,yan2024street} for 3DGS extend these two methods to driving scenes.
However, in driving scenes, all images are captured along recorded trajectories. These methods cannot render high-fidelity images from viewpoints beyond the recorded trajectory due to the lack of supervision on these off-trajectory viewpoints.
\vspace{-4mm}
\paragraph{Off-trajectory view synthesis in driving scenes.} 
Diffusion-based generative methods\cite{rombach2022high,ramesh2022hierarchical,baldridge2024imagen, videoworldsimulators2024, he2022lvdm, blattmann2023stable} have demonstrated notable success in 2D image and video generation.
Inspired by these techniques, some street scene reconstruction methods \cite{yu2024sgd,wang2024freevs,zhao2024drivedreamer4d} employ diffusion models to synthesize off-trajectory views, addressing the challenge of missing views from unrecorded trajectories. For instance, SGD \cite{yu2024sgd} generates novel views using a diffusion process based on reference images and depth maps of the target view. However, it is mainly constrained to synthesizing camera views with only rotation changes and may struggle with large spatial translation.
FreeVS \cite{wang2024freevs} uses a pseudo image of the target view as a prior for the diffusion model. Nevertheless, this pseudo image is derived from the LiDAR point cloud, which limits the model's ability to generate regions unreachable by LiDAR.
In addition to the diffusion model, UniSim \cite{yang2023unisim} adopted GAN-based~\cite{goodfellow2020generative} supervision to enhance photorealism in off-trajectory views.
However, compared with state-of-the-art diffusion models, GAN limits photorealism in largely deviated viewpoints.
AutoSplat~\cite{khan2024autosplat} leverages the symmetry prior of 3D vehicle shape templates as constraints to improve quality after viewpoint deviations.

\begin{figure*}[!t]
    \centering
   \includegraphics[width=\linewidth]{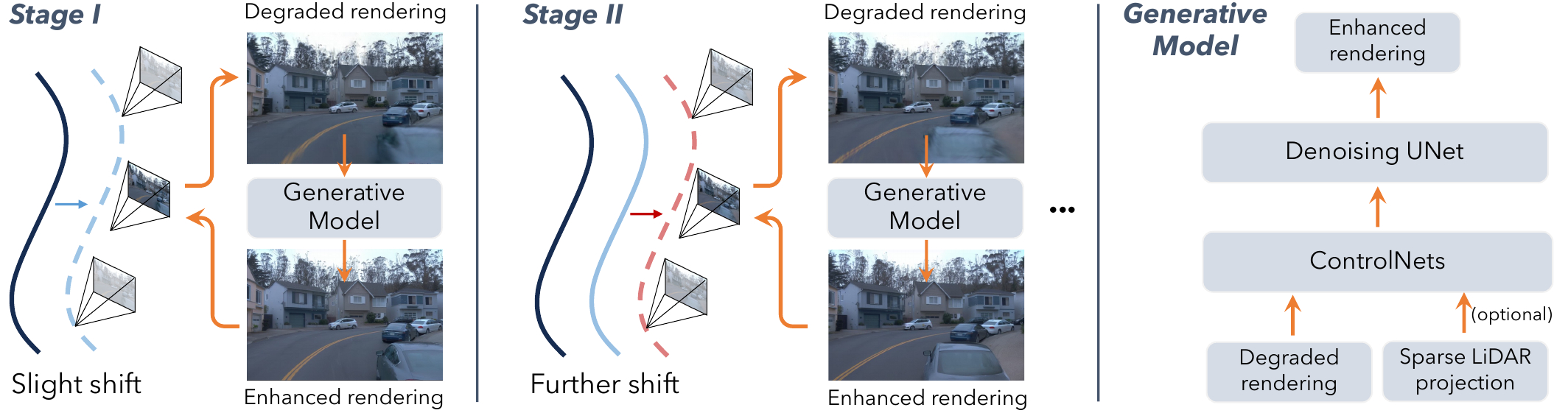}
    \caption{\textbf{The overall framework of \textit{FreeSim}}. 
    The black line indicates the recorded trajectory. Starting from it, the training viewpoints are progressively expanded to far away unrecorded trajectories (\textcolor{cyan}{blue} and \red{red} lines). The generative model produces high-quality images for the new viewpoints after each expansion.
    The progressive expansion can be conducted repeatedly, and we only illustrate two stages for simplicity.
    We use \textbf{real images} from our experiments in this illustration, so readers can zoom in to see the evolution of the four images for an intuitive understanding.
    } 
    \vspace{-3mm}
    \label{fig:framework}
\end{figure*}
\section{Method}

The proposed \name is a generation-reconstruction hybrid method, as demonstrated by \cref{fig:framework}. Here we briefly present the overview from the aspects of generation and reconstruction, respectively.
\vspace{-3mm}
\paragraph{Generation part.}
As discussed in \cref{sec:intro}, we formulate the pose-conditioned view generation task into a generative image enhancement task.
This formulation avoids directly using multi-view extrinsic transformation as conditions, which is unavailable in the single-pass driving trajectories.
For training of the generation model, we first propose a data construction strategy to create training data, presented in \cref{sec:data_construction}.
We then present the model structures and training scheme in ~\cref{sec:model_structure}.
\vspace{-3mm}
\paragraph{Reconstruction part.}
It could be expected that the generative model might struggle to generate high-quality results if the rendering is extremely degraded with large viewpoint changes.
Thus we propose a progressive reconstruction strategy, which progressively adds generated viewpoints into the training image set for reconstruction, starting from those closest to the recorded trajectory and moving progressively farther away.
This part is presented in \cref{sec:recon}.

\subsection{Training Data Construction for Generation}
\label{sec:data_construction}
\paragraph{Data scarcity in our task.} 
Recently, there has been impressive progress in generating multiview images with diffusion models~\cite{gao2024cat3d, voleti2025sv3d, yu2024viewcrafter, liu20243dgs}.
However, there is an essential difference between their paradigms and our task.
These multi-view generative models are trained with multiview samples in NVS datasets such as Objaverse~\cite{deitke2023objaverse}, RealEstate10k~\cite{zhou2018stereo}, CO3D~\cite{reizenstein2021common}, DL3DV~\cite{Ling_2024_CVPR} and  MVImgNet~\cite{yu2023mvimgnet}.
In contrast, the ground-truth images beyond the recorded trajectories in our task are unavailable since the vehicle cannot drive along multiple trajectories simultaneously.
Such data scarcity is our core challenge as well as the essential motivations of our proposed method.

\paragraph{What kind of training data do we need?} To tackle the challenge of data scarcity, we do not train the model to generate off-trajectory views directly from the pose transformation.
Instead, we formulate the view generation task as an image enhancement task.
Specifically, given a radiance field reconstructed from recorded views, we can obtain \emph{\textbf{degraded renderings}} in slightly deviated viewpoints (e.g., laterally shift half meters).
Although the degraded images may be of low quality, they serve as a strong prior to generating the high-quality images.
This formulation allows us to create the training pairs (low-quality and high-quality pairs) based on the recorded views, sidestepping the unreachable requirements of off-trajectory ground truth.
We introduce the way to create such training pairs in the following.

\subsubsection{Preparing Degraded Renderings}

\label{sec:degraded_rendering}

We need to address two issues to obtain proper degraded renderings.
(1) How to efficiently reconstruct numerous scenes to get sufficient data.
(2) How to narrow the gap between the degraded patterns of recorded viewpoints (for training) and the unrecorded viewpoints (for inference).
For the first issue, we propose \emph{Piece-wise Gaussian Reconstruction}.
For the second issue, we adopt \emph{extrapolated rendering} and \emph{Gaussian perturbation} techniques.

\begin{figure} 
    \centering
    \includegraphics[width=0.98\linewidth]{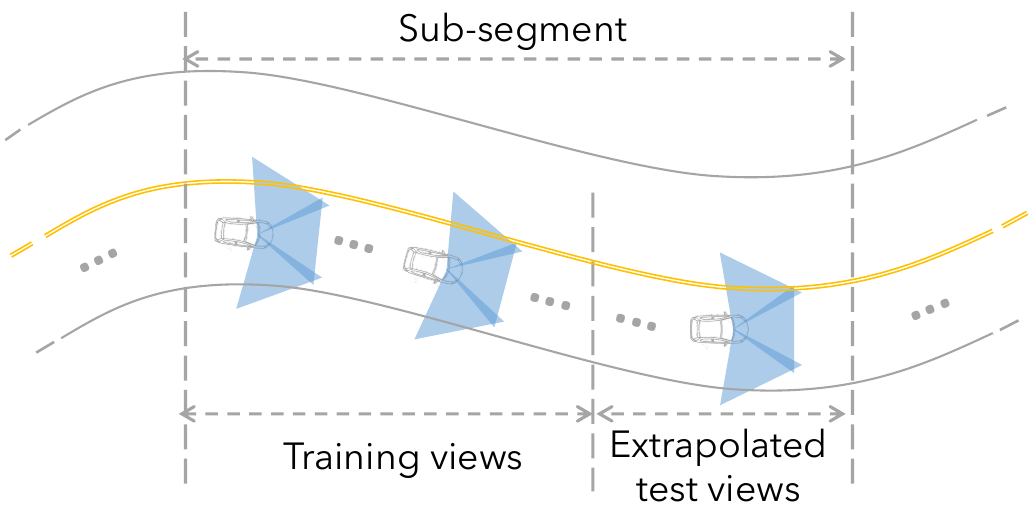}
    \caption{\textbf{Piece-wise Gaussian field reconstruction}.}
    \vspace{-4mm}
    \label{fig:clipped_gs}
\end{figure}

\paragraph{Piece-wise Gaussian reconstruction.}
We reconstruct the Waymo Open Dataset (WOD) ~\cite{Sun_2020_CVPR} to obtain the degraded rendering images.
However, as one of the largest datasets in driving scenes, it contains 1150 block-scale scenes.
It is quite expensive to reconstruct the full dataset.
To make it more efficient, we break each full trajectory into a couple of small sub-segments and reconstruct a small-scale piece-wise Gaussian field for each sub-segment, demonstrated in \cref{fig:clipped_gs}.
Because of its small scale, the piece-wise Gaussian field needs much fewer Gaussian primitives and rapidly converges.
We propose several techniques to match the small scale of piece-wise Gaussian fields and make the reconstruction cheaper.
(1) We resize original images to a half size. (2) We adopt a 1k-iteration schedule with a more aggressive learning rate. (3) The maximum number of Gaussian primitives is set to 1M.
Since our goal here is obtaining degraded renderings, we do not necessitate high-quality reconstructions and are safe to adopt these techniques.
We use the pioneering driving reconstruction method PVG~\cite{chen2023periodic} as the reconstruction tool due to its simplicity.
Thanks to the efficiency of Gaussian-based PVG and our techniques, we could reconstruct a sub-segment in less than 2 minutes.
The whole reconstruction of WOD takes around 40 hours using 8 GPUs, achieving around 6 $\times$ acceleration than full-segment reconstruction.
\par
In addition to efficiency, the piece-wise Gaussian reconstruction further enables better mimics of degraded patterns, presented as follows.

\vspace{-4mm}
\paragraph{Degraded patterns from extrapolated rendering.}
The conventional NVS methods sample test views in an \emph{interpolated} manner.
Unlike them, for each sub-segment, we hold out the last few frames in the clip as test views, which is an \emph{extrapolated} manner.
We adopt this design because our target views, namely the off-trajectory views, are also essentially \emph{extrapolated} views. 
Our design has a more intuitive explanation if we take a side-view camera as an example.
The movement of the side-view camera along the moving-forward trajectory is almost equivalent to the lateral shift for the front cameras, in terms of motion patterns.

\begin{figure}[h]
    \centering
   \includegraphics[width=\linewidth]{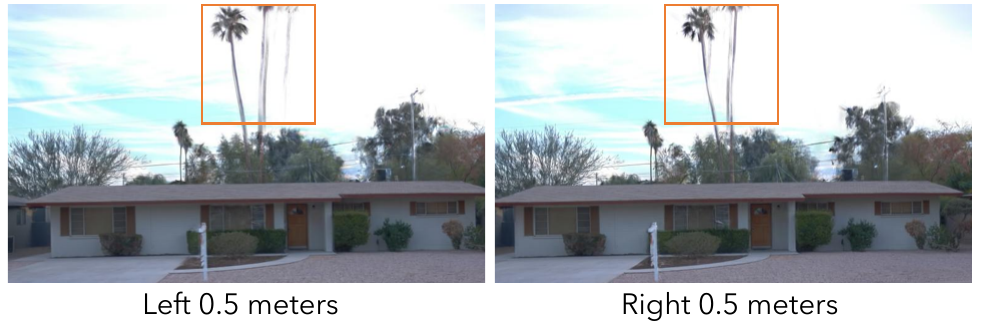}
   \vspace{-4mm}
    \caption{\textbf{Object ghosting caused by inaccurate depth}. The two images are rendered by the original PVG.}
    \vspace{-4mm}
    \label{fig:ghost}
\end{figure}

\vspace{-3mm}
\paragraph{Degraded patterns from Gaussian perturbation.}
To increase the data diversity, we further create the degraded renderings by adding noise to Gaussian primitives in the piece-wise Gaussian field. 
A very typical degraded pattern after viewpoint deviation is ``object ghosting'' illustrated in \cref{fig:ghost}.
This pattern is caused by those Gaussian primitives with inaccurate depth, which is rasterized into wrong image positions after the viewpoint deviations.
To simulate the ghosting, we randomly sample a small portion of Gaussian primitives in a scene and move them the same random distance along the horizontal direction (i.e., width direction in image space).
Thus, the perturbed Gaussian primitives produce ghosting after rasterization.
Moreover, the relative rotations of primitives to the camera also change after a viewpoint deviation.
Thus we further add minor noise to the rotation of Gaussian primitives in their canonical coordinates to better simulate degraded renderings.

\vspace{-4mm}
\paragraph{Image blending.}
A potential concern is that the renderings created by our strategy can be over-degraded, causing the inability of the trained model to handle slight degradation.
To this end, we adopt an image blending strategy, formulated as $\mathbf{I}^\prime = \alpha \mathbf{I}_d + (1 - \alpha)\mathbf{I}^\ast$, which blends degraded image $\mathbf{I}_d$ with ground-truth image $\mathbf{I}^\ast$. Note this process is conducted during training, not for data construction.

\vspace{-4mm}
\paragraph{Auxiliary sparse LiDAR condition. }
In addition to the degraded images, we optionally incorporate LiDAR conditions proposed in FreeVS~\cite{wang2024freevs}.
Specifically, the 3D point clouds are projected into the image plane, resulting in a pseudo image. Note that the point clouds can be assigned with color information by calibration with recorded images.

The sparse LiDAR input serves as a good auxiliary input in our setting, providing more accurate geometry, especially for those nearby regions with small depths (i.e., large disparities).
Moreover, the limitations of LiDAR conditions, such as insufficient coverage and sparsity, can be effectively addressed by our dense rendering conditions.


\subsubsection{Summary of Constructed Training Data}
Finally, we obtain training samples formulated as 
\begin{equation}
    \{\mathbf{I}_d, \, \mathbf{I}_l, \, \mathbf{I}^\ast\}.
    \label{eq:data_sample}
\end{equation}
In \cref{eq:data_sample}, $\mathbf{I}_d$ is a degraded image created by extrapolated rendering or Gaussian perturbation.
$\mathbf{I}_l$ is the LiDAR condition.
$\mathbf{I}^\ast$ is the corresponding recorded ground-truth image.
For each recorded image, we create multiple training pairs using different degradation types or noise parameters.
In total, we collect around 1.5M training pairs.
\subsection{Structure of Generative Model}
\label{sec:model_structure}

We adopt the popular Stable Diffusion (SD) v1.5 \cite{rombach2022high} as our base model.
To leverage the conditional signal from degraded renderings and the LiDAR projection, we integrate two ControlNets \cite{zhang2023adding} into the SD 1.5 backbone, for image condition $\mathbf{I}_d$ and LiDAR condition $\mathbf{I}_l$, respectively.
The resulting features from the two ControlNet are added together and fused into the blocks at each resolution level of the denoising UNet.
The training Loss $\mathcal{L}$ can be written as 

\begin{equation}
    \mathcal{L} = \mathbb{E}_{z_0,c_d,c_l,t,\epsilon} \left[\parallel {\epsilon_\theta}(z_t;c_d,c_l,t) -\epsilon \parallel^2_2 \right]
    \label{eq:Diffusion_Loss}
\end{equation}

where $z_0$ represents the latent of a ground truth image $\mathbf{I}^\ast$ encoded by a VAE~\cite{kingma2013auto}.
$\epsilon_\theta$ is our trainable denoising model.
$z_t$ denotes the noisy latents with denoising time step $t$. $c_d$ and $c_l$ are the latents of image condition $\mathbf{I}_d$ and LiDAR condition $\mathbf{I}_l$, respectively, encoded by a tiny encoder presented in ControlNet.
The latents $c_d$ and $c_l$ are then processed through their respective ControlNet branches.
We remove the cross-attention associated with CLIP~\cite{radford2021learning} text embeddings.
For other settings, we adopt the same configurations as those used in Stable Diffusion 1.5 and ControlNet.

\subsection{Reconstruction}
\label{sec:recon}
After finishing the training of the generation model, we move on to the reconstruction part.
\vspace{-3mm}
\paragraph{Gaussian field initialization.}
Since the generation model is based on degraded renderings, we conduct a pre-reconstruction using the standard procedure in PVG~\cite{chen2023periodic} to offer a relatively good initial status. Afterward, the off-trajectory viewpoints join the game as follows.

\vspace{-3mm}
\paragraph{Progressive viewpoint expansion.}
Although the generative model could enhance the degraded renderings, it can be expected that the quality of generation might be insufficient if the rendering is completely ruined in a largely deviated viewpoint.
To avoid this issue, we propose to incorporate off-trajectory viewpoints progressively, starting from small viewpoint deviations to large viewpoint deviations. 

There are various orders for progressively incorporating off-trajectory viewpoints, here we adopt laterally trajectory shifting as demonstrated in \cref{fig:framework}.
This lane-change-like lateral shifting is the most common yet also the most challenging simulation requirement.
Specifically, for the first update, we laterally shift all viewpoints in the trajectory by a pre-defined step size and generate new images in the shifted viewpoints.
Then generated images are added to the training set.
After each training set update, we freeze the training set and optimize the Gaussian field until it almost converges.
Afterward, we shift the last updated trajectory again and repeat the generation-reconstruction process.
Such a progressive manner ensures the renderings of newly added viewpoints always degrade slightly and the generation model can easily recover them.

 \vspace{-3mm}
\paragraph{Post-enhancement to mitigate rolling shutter distortion and generative randomness.}
In the reconstruction process, there are two issues that reduce the final reconstruction quality: the well-known rolling shutter distortion and generative randomness. 
The cameras in driving scenes are usually moving at a non-negligible speed during recording, leading to rolling shutter distortion in the captured images.
Thus the underlying 3D structures are also distorted with the standard pinhole camera model in 3DGS and NeRF.
In our case, this issue gets even worse since we incorporate more viewpoints than the single trajectory reconstruction.
Under this circumstance, the underlying 3D structure that fits every (distorted) view may not even exist.
Additionally, the generative model inevitably introduces randomness, such as slightly changing the detailed textures, leading to inconsistency between multiple viewpoints.
\par
These two issues cause a slight blur in the rendering results after our progressive reconstruction.
To further increase the quality, we apply our generative enhancement model to the rendering results again as an effective post-processing.

\begin{figure*}[!t]
    \centering
    \includegraphics[width=\linewidth]{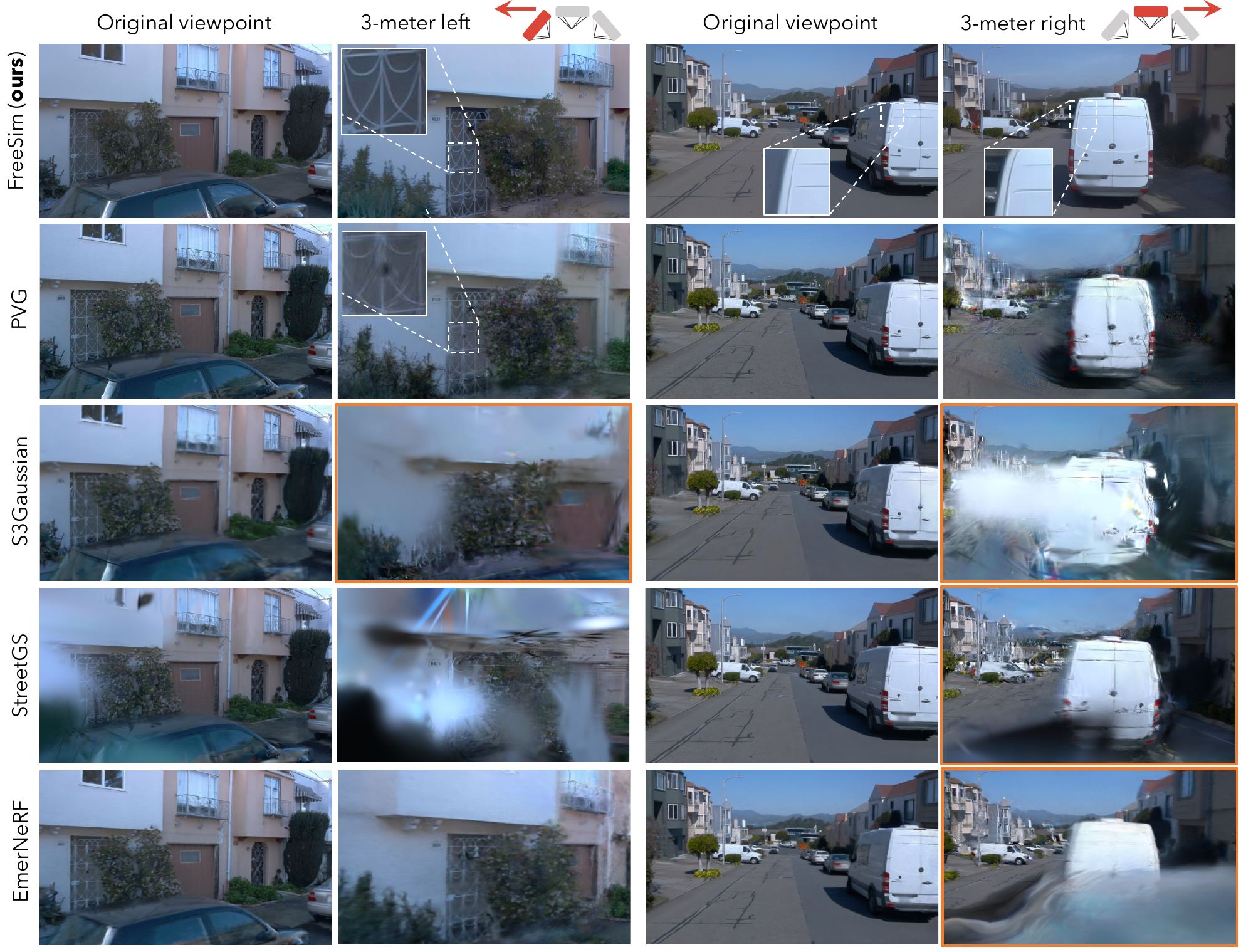}
    \caption{\textbf{Qualitative comparison}. The two scenes are captured by the front-left camera and the front camera, respectively.
    To showcase our performance, we adopt a quite large viewpoint change. However, those views with \textcolor{orange}{orange} bounding boxes get completely ruined with such a large deviation. Thus, we slightly adjust their camera poses.}
    \label{fig:Qualitative_comparison}
\end{figure*}
\section{Experiments}

\subsection{Setup}
\paragraph{Dataset.}
We conduct experiments in the large-scale Waymo Open Dataset~\cite{Sun_2020_CVPR} (WOD), which effectively meets our data requirements for the training of generation.
We select 16 driving sequences from the candidate scenes adopted in EmerNeRF~\cite{yang2023emernerf} 
for the reconstruction part. 
Since we need to shift the driving trajectories, we choose these 16 scenes with relatively large moving space, avoiding moving the cameras to the inside space of obstacles. 
For each driving sequence, we directly adopt the first 50 frames without selection to reconstruct the scenes for simplicity. 
\vspace{-3mm}
\paragraph{Details of models and training.}
Our generative model is initialized from the pretrained weights of Stable Diffusion 1.5. We train all the model's parameters for 50k iterations in 8 A100 GPUs with a batch size of 12 for each GPU. The image resolution is set to $960 \times 640$, which is the half size of the original resolution in WOD.
The $\alpha$ in image blending is set to 0.5, and it is enabled with a probability of 0.1 during training.
For the reconstruction part, we use a pretrained PVG model as Gaussian field initialization to ensure our enhancement model could work at the early training stage.
We further conduct the optimization for another 30k steps along with viewpoints expansion.
By default, we conduct a viewpoint expansion every 5k iterations with a step size of 0.5 meters.
Other settings follow the default PVG~\cite{chen2023periodic}.

\vspace{-3mm}
\paragraph{Details of data construction.}
For data construction, we reconstruct the whole training split of WOD with the proposed techniques in \cref{sec:degraded_rendering}, using PVG as the reconstruction tool.
The length of sub-segments is 20 frames.
For each one, we adopt the last 4 frames for extrapolated rendering.
We also render all the training views with the Gaussian perturbation technique.
For the perturbation, we randomly sample at most 50\% Gaussian primitives to add noise.
The max translation noise is 0.2 meters and the max rotation noise is 15 degrees.
In total, we create around 1.5M samples, taking around 40 hours with 8 GPUs.

\begin{figure*}[t]
    \centering
    \includegraphics[width=\linewidth]{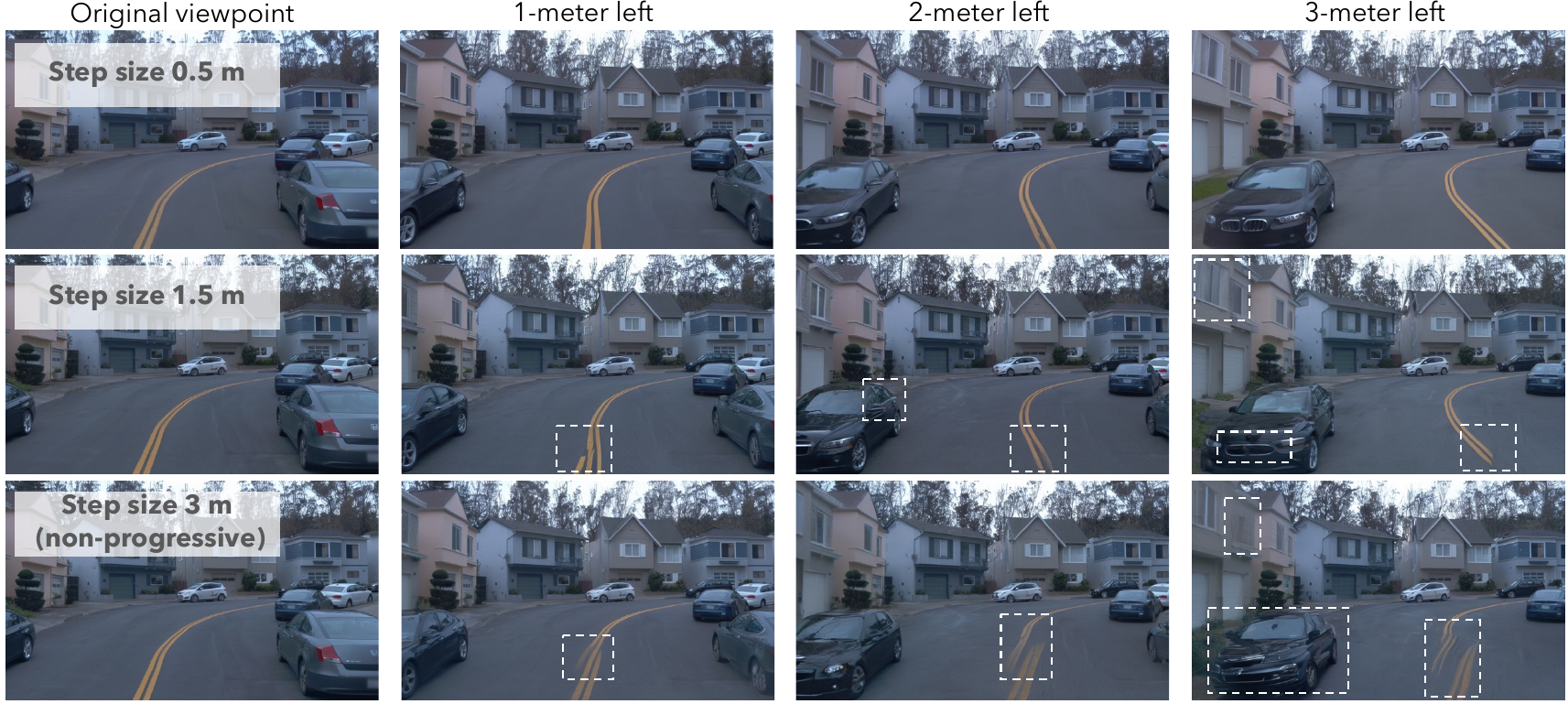}
    \vspace{-5mm}
    \caption{\textbf{Demonstration of different step sizes in progressive reconstruction}. All images are enhanced by the proposed post-enhancement. Empirically, step sizes less than 1 meter usually produce good results.  }
    \vspace{-3mm}
    \label{fig:progressive}
\end{figure*}

\subsection{Main results}
\paragraph{Quantitative comparison.}

As mentioned, there are no ground truth images in the off-trajectory viewpoints. 
Thus, following previous approaches~\cite{yu2024sgd, yang2023unisim, wang2024freevs}, we adopt FID score for off-trajectory evaluation.
This score evaluates the distributional similarity between off-trajectory rendering images and the original ground truth images from the recorded trajectories.
As shown in \cref{tab:main_results}, the proposed \name significantly outperforms previous methods.
We also report the conventional on-trajectory Novel View Synthesis (NVS) metrics PSNR and SSIM following EmerNeRF~\cite{yang2023emernerf}.
In the recorded trajectories, our method has a slight performance drop. This is mainly because the generation part inevitably introduces minor inconsistencies. However, we emphasize that the visual quality of our results remains good and is sufficient for simulation purposes.

\begin{table}[h]
\begin{center}
\resizebox{\columnwidth}{!}{
\begin{tabular}{l|cc|ccc}
\toprule
\multirow{2}{*}{Method} &
\multicolumn{2}{c|}{Recorded views} & 
\multicolumn{3}{c}{Unrecorded views (FID $\downarrow$)} \\
& PSNR $\uparrow$ & SSIM $\uparrow$  & @1m & @2m & @3m \\
\midrule
StreetGS~\cite{yan2024street} & 28.01 & {\bf0.871} & 25.8 & 35.4 & 47.6 \\
EmerNeRF~\cite{yang2023emernerf} & 29.18 & 0.842 & 32.3 & 40.2 & 49.8\\
S3Gaussian~\cite{huang2024s3gaussian} & 26.84 & 0.844 & 59.4 & 76.4 & 81.3 \\
PVG (baseline)~\cite{chen2023periodic} & {\bf29.19} & 0.866  & 22.9 & 34.3 & 47.5 \\
FreeSim (ours) & 28.32 & 0.852  & {\bf14.6} & {\bf17.0} & {\bf18.6} \\

\bottomrule
\end{tabular}}
\end{center}
\vspace{-4mm}
\caption{\textbf{Quantitative comparison}. We report both the performance of novel view synthesis on recorded trajectories and the performance on off-trajectory viewpoints, with lateral trajectory shifts of 1m, 2m, and 3m.}
\vspace{-4mm}
\label{tab:main_results}
\end{table}

\paragraph{Qualitative comparison.}
\cref{fig:Qualitative_comparison} shows the visual comparison between FreeSim and other methods in original views and off-trajectory views.

\vspace{-3mm}
\paragraph{Discussion with related work.}
There are a few concurrent preprinted works sharing a similar goal with us, including FreeVS~\cite{wang2024freevs} and DriveDreamer4D~\cite{zhao2024drivedreamer4d}.
Our method has some essential differences compared with them.
FreeVS is a pure generative method conditioned on LiDAR projection, which means it is limited by the LiDAR coverage.
DriveDreamer4D~ is built upon generative driving world models~\cite{wang2023drivedreamer,wang2024driving}, which are learned from real driving records.
Thus, it has potential limitations to generate unusual handmade trajectories such as the viewpoint lifting demonstrated in our video.

\subsection{Ablations and Analysis}

\paragraph{Progressive reconstruction.}
\cref{tab:ablation} and \cref{fig:progressive} showcase the quantitative and qualitative ablations, respectively, demonstrating the effectiveness of progressive reconstruction with different step sizes.
In addition, we have further findings as follows.
\begin{enumerate}[leftmargin=*]
    \item Smaller step size leads to slightly lower rendering quality for recorded views. This is because the generation model cannot be perfect and using more generated images for training inevitably introduces more inconsistency. However, our on-trajectory rendering quality is still competitive compared with all methods.
    \item \cref{fig:progressive} shows that larger step sizes cause more artifacts, especially for the regions close to cameras. This is because these regions have large disparities.
\end{enumerate}

\begin{figure}[!h]
    \centering
    \includegraphics[width=\linewidth]{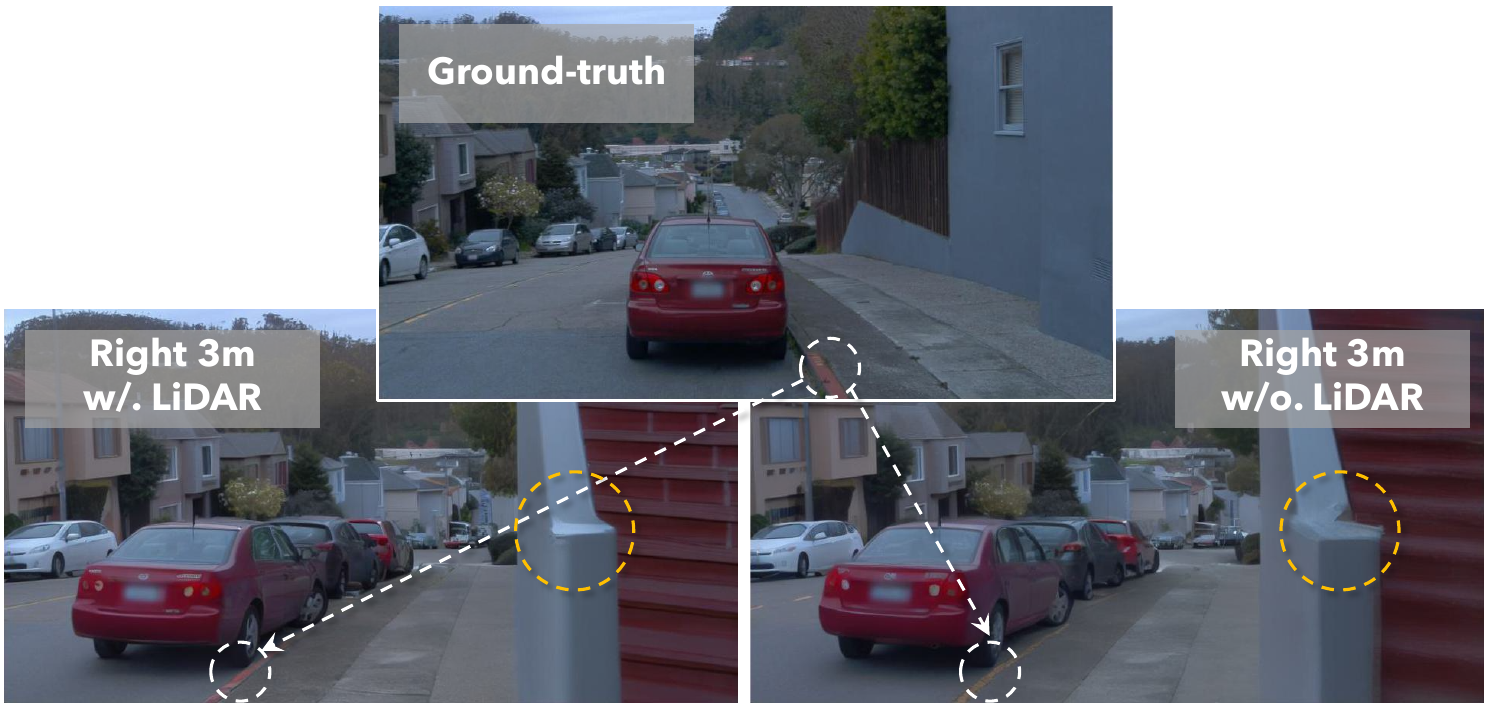}
    \caption{\textbf{LiDAR condition improves robustness in extreme viewpoints.} Here the deviated viewpoint is about to collide with the stairs. In such an extreme view, LiDAR input corrects the color of lines, eliminates the artifact, and makes the stairs more clear.  }
    \vspace{-3mm}
    \label{fig:lidar_demo}
\end{figure}

\vspace{-3mm}
\paragraph{LiDAR condition.}
As demonstrated by \cref{tab:ablation}, disabling the LiDAR condition leads to an acceptable performance drop.
We further find that LiDAR information has more impact on the large deviations as the FID score at 3 meters shows, which is also confirmed by \cref{fig:lidar_demo}.

\begin{table}[h]
\begin{center}
\resizebox{\columnwidth}{!}{
\begin{tabular}{l|cc|ccc}
\toprule
 \multirow{2}{*}{Models} &
\multicolumn{2}{c|}{Recorded views} & 
\multicolumn{3}{c}{Unrecorded views (FID $\downarrow$)} \\
 & PSNR $\uparrow$ & SSIM $\uparrow$ & @1m & @2m & @3m \\
\midrule
\midrule
w/o. LiDAR & 28.61 & 0.854  & 15.5 & 18.5 & 21.3 \\
w/o. Perturb. $\dag$ & 28.32 & 0.851  & 15.4 & 17.1 & 18.9 \\
w/o. Ext. render $\dag$ & 28.25 & 0.851 & 14.5 & 18.3 & 20.3 \\
Blending p = 0.5 $\ast$ & 28.39 & 0.852 & 15.1 & 19.0 & 20.4\\
Blending p = 0.0 & 28.27 & 0.850 & 15.3 & 18.7 & 20.6\\
\midrule 
Step size 0.25 m & 28.12 & 0.846 & 15.6 & 17.7 & 19.5\\ 
Step size 1.0 m & 28.77 & 0.860 & 14.5 & 16.9 & 18.4\\
Step size 1.5 m & 28.93 & 0.863 & 16.0 & 18.2 & 20.0 \\ 
Non-progressive & 29.01 & 0.864 & 20.1 & 26.3 & 29.7 \\
\midrule
Default model & 28.32 & 0.852 & 14.6 & 17.0 & 18.6 \\
\bottomrule
\end{tabular}}
\end{center}
\vspace{-3mm}
\caption{\textbf{Quantitative ablation.} $\dag$: ``Perturb.'' and ``Ext. render'' means the Gaussian perturbation and extrapolated rendering to create degraded patterns in \cref{sec:degraded_rendering}.
$\ast$: ``Blending p'' means the probability of enabling the image blending technique. 
Here the ``Default model'' has a blending p of 0.1 and a step size of 0.5 m.
}
\label{tab:ablation}
\end{table}

\vspace{-4mm}
\paragraph{Ways to make degraded rendering.}
We propose extrapolated rendering and Gaussian perturbation to create degraded renderings.
\cref{tab:ablation} shows we can also obtain relatively good results solely using one of them, while the combination of them leads to better results, especially for large deviations.
Note we cannot disable both of them. If doing so and using LiDAR-only condition, the LiDAR-uncovered regions exhibit randomly generated content and the reconstruction cannot converge.

\vspace{-3mm}
\paragraph{Image blending.}
\cref{tab:ablation} demonstrates that proper image blending (the default model, p = 0.1) produces better results.
However, over-blending (p = 0.5) makes the learning task too simple to learn good representation. 

\begin{figure}[!h]
    \centering
    \includegraphics[width=\linewidth]{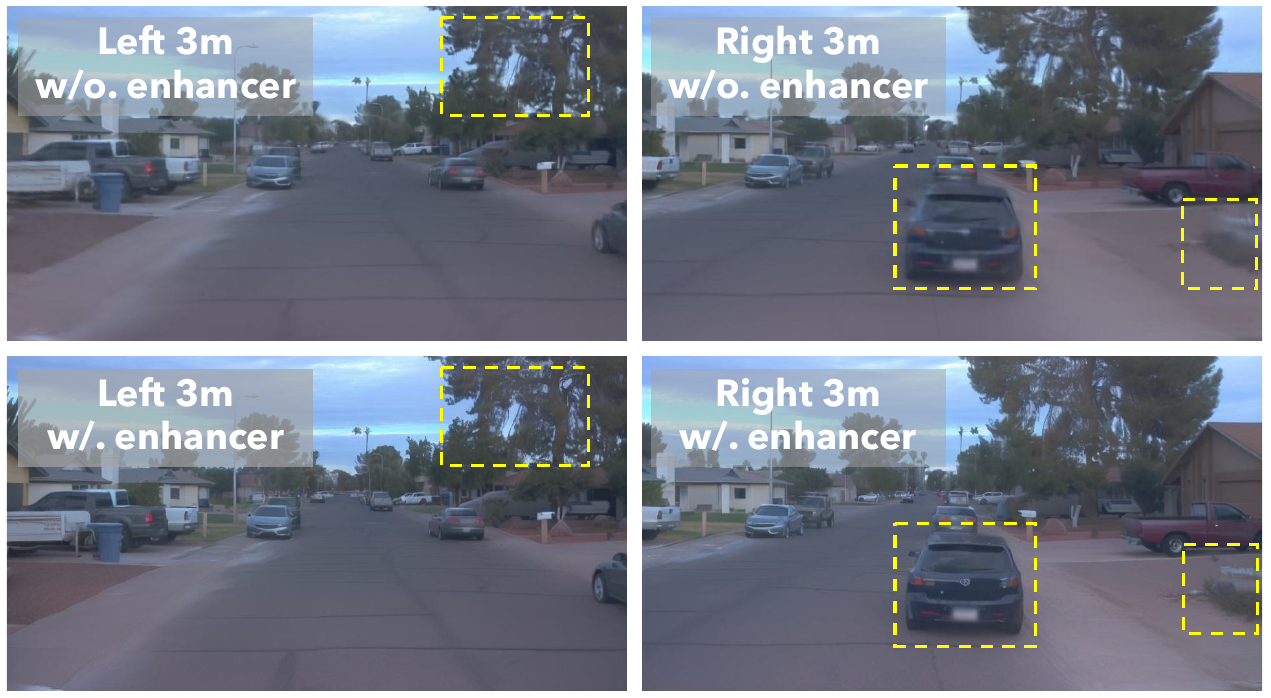}
    \vspace{-4mm}
    \caption{\textbf{Effectiveness of post enhancement.} Post enhancement enhances the detailed texture such as \emph{tree} and \emph{grass} and objects near the camera such as the car.}
    \label{fig:enhancer}
\end{figure}

\vspace{-3mm}
\paragraph{Post enhancement.}
As discussed in \cref{sec:recon}, rolling shutter distortions and generative randomness cause slight blur in the final rendering results.
We propose to apply the enhancement model to the final renderings to increase the quality, demonstrated in \cref{fig:enhancer}.
This technique improves the image quality, especially for high-frequency regions (e.g., \emph{tree} and \emph{grass}) and nearby objects such as the car. 

\begin{figure}[ht]
    \centering
    \includegraphics[width=\linewidth]{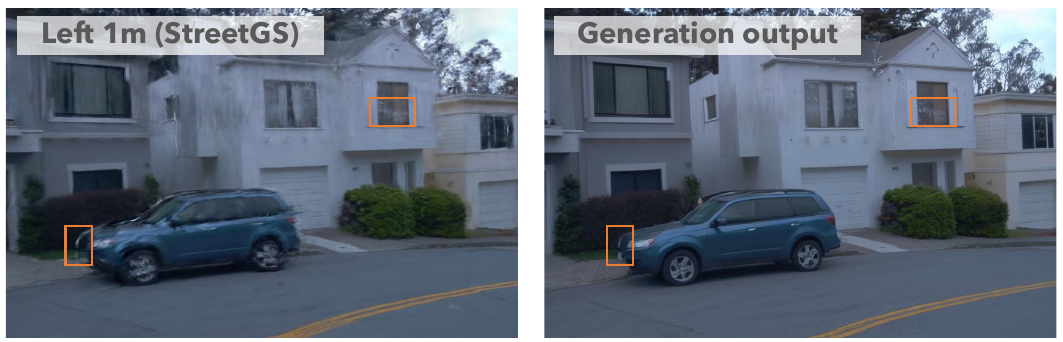}
    \caption{\textbf{The generalization of our generative model.} Although trained with data created by PVG~\cite{chen2023periodic}, the model has good generalization in degraded patterns of different reconstruction methods such as StreetGS~\cite{yan2024street}. The remaining artifacts are likely to be mitigated if we create more training data using StreetGS.}
    \vspace{-4mm}
    \label{fig:generalization}
\end{figure}

\vspace{-5mm}
\paragraph{Generalization of generative model.}
Although our training data is solely created by PVG~\cite{chen2023periodic}, we find it can be generalized to the renderings from different reconstruction methods. \cref{fig:generalization} provides an example in the reconstruction of StreetGS~\cite{yan2024street}. 
The quality can be further improved if we adopt more training data created from StreetGS.

\begin{figure}[ht]
    \centering
    \includegraphics[width=\linewidth]{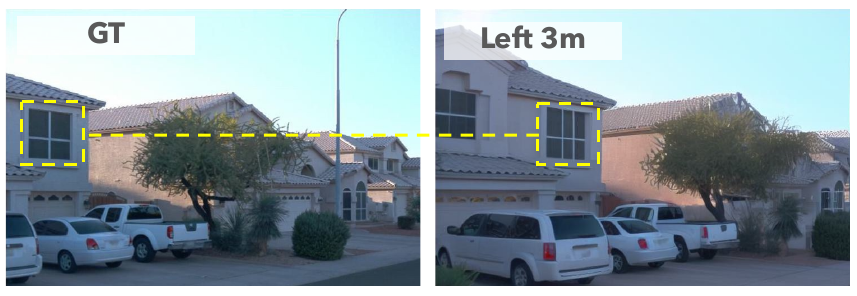}
    \vspace{-4mm}
    \caption{\textbf{A typical failure case}. During the progressive reconstruction, the window frame converges to a different shape, caused by a severe ``ghosting'' issue (\cref{fig:ghost}) after viewpoint deviations.}
    \vspace{-3mm}
    \label{fig:failure}
\end{figure}

\vspace{-3mm}
\paragraph{Failure cases.}
\cref{fig:failure} shows a typical failure case. It is caused by the inaccurate depth of the window frames, making the window gradually converge to a new shape during our progressive reconstruction. 

\section{Conclusion, limitations, and future work}
The proposed \name is a generation-reconstruction hybrid method capable of synthesizing high-quality views largely deviated from the recorded driving trajectories.
However, our method still has several limitations. 

\name cannot perfectly handle the rolling shutter distortion and the randomness of the generation part, leading to blur in regions with rich details such as grass. Although we can improve the quality by post-enhancement, it will reduce the efficiency due to multi-step denoising.

The training data is solely created by PVG, which is a Gaussian-based method. Thus, our generation model cannot generalize well to the NeRF-based methods. Although further adopting NeRFs in the data construction could address this limitation, it may slow the data construction pipeline due to the relatively low efficiency of NeRFs.

Our future work includes adopting video generation models, making the scene editable, and improving the native 3DGS rasterizer to handle the rolling shutter issue.

{
    \small
    \bibliographystyle{ieeenat_fullname}
    \bibliography{main}

\begin{thebibliography}{44}
\providecommand{\natexlab}[1]{#1}
\providecommand{\url}[1]{\texttt{#1}}
\expandafter\ifx\csname urlstyle\endcsname\relax
  \providecommand{\doi}[1]{doi: #1}\else
  \providecommand{\doi}{doi: \begingroup \urlstyle{rm}\Url}\fi

\bibitem[Baldridge et~al.(2024)Baldridge, Bauer, Bhutani, Brichtova, Bunner, Chan, Chen, Dieleman, Du, Eaton-Rosen, et~al.]{baldridge2024imagen}
Jason Baldridge, Jakob Bauer, Mukul Bhutani, Nicole Brichtova, Andrew Bunner, Kelvin Chan, Yichang Chen, Sander Dieleman, Yuqing Du, Zach Eaton-Rosen, et~al.
\newblock Imagen 3.
\newblock \emph{arXiv preprint arXiv:2408.07009}, 2024.

\bibitem[Blattmann et~al.(2023)Blattmann, Dockhorn, Kulal, Mendelevitch, Kilian, Lorenz, Levi, English, Voleti, Letts, et~al.]{blattmann2023stable}
Andreas Blattmann, Tim Dockhorn, Sumith Kulal, Daniel Mendelevitch, Maciej Kilian, Dominik Lorenz, Yam Levi, Zion English, Vikram Voleti, Adam Letts, et~al.
\newblock Stable video diffusion: Scaling latent video diffusion models to large datasets.
\newblock \emph{arXiv preprint arXiv:2311.15127}, 2023.

\bibitem[Brooks et~al.(2024)Brooks, Peebles, Holmes, DePue, Guo, Jing, Schnurr, Taylor, Luhman, Luhman, Ng, Wang, and Ramesh]{videoworldsimulators2024}
Tim Brooks, Bill Peebles, Connor Holmes, Will DePue, Yufei Guo, Li Jing, David Schnurr, Joe Taylor, Troy Luhman, Eric Luhman, Clarence Ng, Ricky Wang, and Aditya Ramesh.
\newblock Video generation models as world simulators.
\newblock 2024.

\bibitem[Chen et~al.(2023)Chen, Gu, Jiang, Zhu, and Zhang]{chen2023periodic}
Yurui Chen, Chun Gu, Junzhe Jiang, Xiatian Zhu, and Li Zhang.
\newblock Periodic vibration gaussian: Dynamic urban scene reconstruction and real-time rendering.
\newblock \emph{arXiv:2311.18561}, 2023.

\bibitem[Chen et~al.(2024)Chen, Yang, Huang, de~Lutio, Esturo, Ivanovic, Litany, Gojcic, Fidler, Pavone, et~al.]{chen2024omnire}
Ziyu Chen, Jiawei Yang, Jiahui Huang, Riccardo de Lutio, Janick~Martinez Esturo, Boris Ivanovic, Or Litany, Zan Gojcic, Sanja Fidler, Marco Pavone, et~al.
\newblock Omnire: Omni urban scene reconstruction.
\newblock \emph{arXiv preprint arXiv:2408.16760}, 2024.

\bibitem[Deitke et~al.(2023)Deitke, Schwenk, Salvador, Weihs, Michel, VanderBilt, Schmidt, Ehsani, Kembhavi, and Farhadi]{deitke2023objaverse}
Matt Deitke, Dustin Schwenk, Jordi Salvador, Luca Weihs, Oscar Michel, Eli VanderBilt, Ludwig Schmidt, Kiana Ehsani, Aniruddha Kembhavi, and Ali Farhadi.
\newblock Objaverse: A universe of annotated 3d objects.
\newblock In \emph{Proceedings of the IEEE/CVF Conference on Computer Vision and Pattern Recognition}, pages 13142--13153, 2023.

\bibitem[Gao et~al.(2024)Gao, Holynski, Henzler, Brussee, Martin-Brualla, Srinivasan, Barron, and Poole]{gao2024cat3d}
Ruiqi Gao, Aleksander Holynski, Philipp Henzler, Arthur Brussee, Ricardo Martin-Brualla, Pratul Srinivasan, Jonathan~T Barron, and Ben Poole.
\newblock Cat3d: Create anything in 3d with multi-view diffusion models.
\newblock \emph{arXiv preprint arXiv:2405.10314}, 2024.

\bibitem[Goodfellow et~al.(2020)Goodfellow, Pouget-Abadie, Mirza, Xu, Warde-Farley, Ozair, Courville, and Bengio]{goodfellow2020generative}
Ian Goodfellow, Jean Pouget-Abadie, Mehdi Mirza, Bing Xu, David Warde-Farley, Sherjil Ozair, Aaron Courville, and Yoshua Bengio.
\newblock Generative adversarial networks.
\newblock \emph{Communications of the ACM}, 63\penalty0 (11):\penalty0 139--144, 2020.

\bibitem[He et~al.(2022)He, Yang, Zhang, Shan, and Chen]{he2022lvdm}
Yingqing He, Tianyu Yang, Yong Zhang, Ying Shan, and Qifeng Chen.
\newblock Latent video diffusion models for high-fidelity long video generation.
\newblock 2022.

\bibitem[Huang et~al.(2024)Huang, Wei, Zheng, An, Lu, Zhan, Tomizuka, Keutzer, and Zhang]{huang2024s3gaussian}
Nan Huang, Xiaobao Wei, Wenzhao Zheng, Pengju An, Ming Lu, Wei Zhan, Masayoshi Tomizuka, Kurt Keutzer, and Shanghang Zhang.
\newblock S3gaussian: Self-supervised street gaussians for autonomous driving.
\newblock \emph{arXiv preprint arXiv:2405.20323}, 2024.

\bibitem[Kerbl et~al.(2023)Kerbl, Kopanas, Leimk{\"u}hler, and Drettakis]{kerbl3Dgaussians}
Bernhard Kerbl, Georgios Kopanas, Thomas Leimk{\"u}hler, and George Drettakis.
\newblock 3d gaussian splatting for real-time radiance field rendering.
\newblock \emph{ACM Transactions on Graphics}, 42\penalty0 (4), 2023.

\bibitem[Khan et~al.(2024)Khan, Fazlali, Sharma, Cao, Bai, Ren, and Liu]{khan2024autosplat}
Mustafa Khan, Hamidreza Fazlali, Dhruv Sharma, Tongtong Cao, Dongfeng Bai, Yuan Ren, and Bingbing Liu.
\newblock Autosplat: Constrained gaussian splatting for autonomous driving scene reconstruction.
\newblock \emph{arXiv preprint arXiv:2407.02598}, 2024.

\bibitem[Kingma(2013)]{kingma2013auto}
Diederik~P Kingma.
\newblock Auto-encoding variational bayes.
\newblock \emph{arXiv preprint arXiv:1312.6114}, 2013.

\bibitem[Ling et~al.(2024)Ling, Sheng, Tu, Zhao, Xin, Wan, Yu, Guo, Yu, Lu, Li, Sun, Ashok, Mukherjee, Kang, Kong, Hua, Zhang, Benes, and Bera]{Ling_2024_CVPR}
Lu Ling, Yichen Sheng, Zhi Tu, Wentian Zhao, Cheng Xin, Kun Wan, Lantao Yu, Qianyu Guo, Zixun Yu, Yawen Lu, Xuanmao Li, Xingpeng Sun, Rohan Ashok, Aniruddha Mukherjee, Hao Kang, Xiangrui Kong, Gang Hua, Tianyi Zhang, Bedrich Benes, and Aniket Bera.
\newblock Dl3dv-10k: A large-scale scene dataset for deep learning-based 3d vision.
\newblock In \emph{Proceedings of the IEEE/CVF Conference on Computer Vision and Pattern Recognition (CVPR)}, pages 22160--22169, 2024.

\bibitem[Liu et~al.(2024{\natexlab{a}})Liu, Sun, Wang, Wang, Sun, Ye, Zhang, and Duan]{liu2024reconx}
Fangfu Liu, Wenqiang Sun, Hanyang Wang, Yikai Wang, Haowen Sun, Junliang Ye, Jun Zhang, and Yueqi Duan.
\newblock Reconx: Reconstruct any scene from sparse views with video diffusion model.
\newblock \emph{arXiv preprint arXiv:2408.16767}, 2024{\natexlab{a}}.

\bibitem[Liu et~al.(2023)Liu, Chen, Yang, Wang, Manivasagam, and Urtasun]{liu2023real}
Jeffrey~Yunfan Liu, Yun Chen, Ze Yang, Jingkang Wang, Sivabalan Manivasagam, and Raquel Urtasun.
\newblock Real-time neural rasterization for large scenes.
\newblock In \emph{Proceedings of the IEEE/CVF International Conference on Computer Vision}, pages 8416--8427, 2023.

\bibitem[Liu et~al.(2024{\natexlab{b}})Liu, Zhou, and Huang]{liu20243dgs}
Xi Liu, Chaoyi Zhou, and Siyu Huang.
\newblock 3dgs-enhancer: Enhancing unbounded 3d gaussian splatting with view-consistent 2d diffusion priors.
\newblock \emph{arXiv preprint arXiv:2410.16266}, 2024{\natexlab{b}}.

\bibitem[Luiten et~al.(2023)Luiten, Kopanas, Leibe, and Ramanan]{luiten2023dynamic}
Jonathon Luiten, Georgios Kopanas, Bastian Leibe, and Deva Ramanan.
\newblock Dynamic 3d gaussians: Tracking by persistent dynamic view synthesis.
\newblock \emph{arXiv preprint arXiv:2308.09713}, 2023.

\bibitem[Mildenhall et~al.(2021)Mildenhall, Srinivasan, Tancik, Barron, Ramamoorthi, and Ng]{mildenhall2021nerf}
Ben Mildenhall, Pratul~P Srinivasan, Matthew Tancik, Jonathan~T Barron, Ravi Ramamoorthi, and Ren Ng.
\newblock Nerf: Representing scenes as neural radiance fields for view synthesis.
\newblock \emph{Communications of the ACM}, 65\penalty0 (1):\penalty0 99--106, 2021.

\bibitem[Radford et~al.(2021)Radford, Kim, Hallacy, Ramesh, Goh, Agarwal, Sastry, Askell, Mishkin, Clark, et~al.]{radford2021learning}
Alec Radford, Jong~Wook Kim, Chris Hallacy, Aditya Ramesh, Gabriel Goh, Sandhini Agarwal, Girish Sastry, Amanda Askell, Pamela Mishkin, Jack Clark, et~al.
\newblock Learning transferable visual models from natural language supervision.
\newblock In \emph{International conference on machine learning}, pages 8748--8763. PMLR, 2021.

\bibitem[Ramesh et~al.(2022)Ramesh, Dhariwal, Nichol, Chu, and Chen]{ramesh2022hierarchical}
Aditya Ramesh, Prafulla Dhariwal, Alex Nichol, Casey Chu, and Mark Chen.
\newblock Hierarchical text-conditional image generation with clip latents.
\newblock \emph{arXiv preprint arXiv:2204.06125}, 1\penalty0 (2):\penalty0 3, 2022.

\bibitem[Reizenstein et~al.(2021)Reizenstein, Shapovalov, Henzler, Sbordone, Labatut, and Novotny]{reizenstein2021common}
Jeremy Reizenstein, Roman Shapovalov, Philipp Henzler, Luca Sbordone, Patrick Labatut, and David Novotny.
\newblock Common objects in 3d: Large-scale learning and evaluation of real-life 3d category reconstruction.
\newblock In \emph{Proceedings of the IEEE/CVF international conference on computer vision}, pages 10901--10911, 2021.

\bibitem[Rombach et~al.(2022)Rombach, Blattmann, Lorenz, Esser, and Ommer]{rombach2022high}
Robin Rombach, Andreas Blattmann, Dominik Lorenz, Patrick Esser, and Bj{\"o}rn Ommer.
\newblock High-resolution image synthesis with latent diffusion models.
\newblock In \emph{Proceedings of the IEEE/CVF conference on computer vision and pattern recognition}, pages 10684--10695, 2022.

\bibitem[Sun et~al.(2020)Sun, Kretzschmar, Dotiwalla, Chouard, Patnaik, Tsui, Guo, Zhou, Chai, Caine, Vasudevan, Han, Ngiam, Zhao, Timofeev, Ettinger, Krivokon, Gao, Joshi, Zhang, Shlens, Chen, and Anguelov]{Sun_2020_CVPR}
Pei Sun, Henrik Kretzschmar, Xerxes Dotiwalla, Aurelien Chouard, Vijaysai Patnaik, Paul Tsui, James Guo, Yin Zhou, Yuning Chai, Benjamin Caine, Vijay Vasudevan, Wei Han, Jiquan Ngiam, Hang Zhao, Aleksei Timofeev, Scott Ettinger, Maxim Krivokon, Amy Gao, Aditya Joshi, Yu Zhang, Jonathon Shlens, Zhifeng Chen, and Dragomir Anguelov.
\newblock Scalability in perception for autonomous driving: Waymo open dataset.
\newblock In \emph{Proceedings of the IEEE/CVF Conference on Computer Vision and Pattern Recognition (CVPR)}, 2020.

\bibitem[Tonderski et~al.(2024)Tonderski, Lindstr{\"o}m, Hess, Ljungbergh, Svensson, and Petersson]{tonderski2024neurad}
Adam Tonderski, Carl Lindstr{\"o}m, Georg Hess, William Ljungbergh, Lennart Svensson, and Christoffer Petersson.
\newblock Neurad: Neural rendering for autonomous driving.
\newblock In \emph{Proceedings of the IEEE/CVF Conference on Computer Vision and Pattern Recognition}, pages 14895--14904, 2024.

\bibitem[Voleti et~al.(2025)Voleti, Yao, Boss, Letts, Pankratz, Tochilkin, Laforte, Rombach, and Jampani]{voleti2025sv3d}
Vikram Voleti, Chun-Han Yao, Mark Boss, Adam Letts, David Pankratz, Dmitry Tochilkin, Christian Laforte, Robin Rombach, and Varun Jampani.
\newblock Sv3d: Novel multi-view synthesis and 3d generation from a single image using latent video diffusion.
\newblock In \emph{European Conference on Computer Vision}, pages 439--457. Springer, 2025.

\bibitem[Wang et~al.(2024{\natexlab{a}})Wang, Fan, Wang, Chen, and Zhang]{wang2024freevs}
Qitai Wang, Lue Fan, Yuqi Wang, Yuntao Chen, and Zhaoxiang Zhang.
\newblock Freevs: Generative view synthesis on free driving trajectory.
\newblock \emph{arXiv preprint arXiv:2410.18079}, 2024{\natexlab{a}}.

\bibitem[Wang et~al.(2023)Wang, Zhu, Huang, Chen, Zhu, and Lu]{wang2023drivedreamer}
Xiaofeng Wang, Zheng Zhu, Guan Huang, Xinze Chen, Jiagang Zhu, and Jiwen Lu.
\newblock Drivedreamer: Towards real-world-driven world models for autonomous driving.
\newblock \emph{arXiv preprint arXiv:2309.09777}, 2023.

\bibitem[Wang et~al.(2024{\natexlab{b}})Wang, He, Fan, Li, Chen, and Zhang]{wang2024driving}
Yuqi Wang, Jiawei He, Lue Fan, Hongxin Li, Yuntao Chen, and Zhaoxiang Zhang.
\newblock Driving into the future: Multiview visual forecasting and planning with world model for autonomous driving.
\newblock In \emph{Proceedings of the IEEE/CVF Conference on Computer Vision and Pattern Recognition}, pages 14749--14759, 2024{\natexlab{b}}.

\bibitem[Wu et~al.(2024)Wu, Yi, Fang, Xie, Zhang, Wei, Liu, Tian, and Wang]{wu20244d}
Guanjun Wu, Taoran Yi, Jiemin Fang, Lingxi Xie, Xiaopeng Zhang, Wei Wei, Wenyu Liu, Qi Tian, and Xinggang Wang.
\newblock 4d gaussian splatting for real-time dynamic scene rendering.
\newblock In \emph{Proceedings of the IEEE/CVF Conference on Computer Vision and Pattern Recognition}, pages 20310--20320, 2024.

\bibitem[Wu et~al.(2023)Wu, Liu, Luo, Zhong, Chen, Xiao, Hou, Lou, Chen, Yang, et~al.]{wu2023mars}
Zirui Wu, Tianyu Liu, Liyi Luo, Zhide Zhong, Jianteng Chen, Hongmin Xiao, Chao Hou, Haozhe Lou, Yuantao Chen, Runyi Yang, et~al.
\newblock Mars: An instance-aware, modular and realistic simulator for autonomous driving.
\newblock In \emph{CAAI International Conference on Artificial Intelligence}, pages 3--15. Springer, 2023.

\bibitem[Yan et~al.(2024)Yan, Lin, Zhou, Wang, Sun, Zhan, Lang, Zhou, and Peng]{yan2024street}
Yunzhi Yan, Haotong Lin, Chenxu Zhou, Weijie Wang, Haiyang Sun, Kun Zhan, Xianpeng Lang, Xiaowei Zhou, and Sida Peng.
\newblock Street gaussians for modeling dynamic urban scenes.
\newblock \emph{arXiv preprint arXiv:2401.01339}, 2024.

\bibitem[Yang et~al.(2023{\natexlab{a}})Yang, Ivanovic, Litany, Weng, Kim, Li, Che, Xu, Fidler, Pavone, and Wang]{yang2023emernerf}
Jiawei Yang, Boris Ivanovic, Or Litany, Xinshuo Weng, Seung~Wook Kim, Boyi Li, Tong Che, Danfei Xu, Sanja Fidler, Marco Pavone, and Yue Wang.
\newblock Emernerf: Emergent spatial-temporal scene decomposition via self-supervision.
\newblock \emph{arXiv preprint arXiv:2311.02077}, 2023{\natexlab{a}}.

\bibitem[Yang et~al.(2023{\natexlab{b}})Yang, Chen, Wang, Manivasagam, Ma, Yang, and Urtasun]{yang2023unisim}
Ze Yang, Yun Chen, Jingkang Wang, Sivabalan Manivasagam, Wei-Chiu Ma, Anqi~Joyce Yang, and Raquel Urtasun.
\newblock Unisim: A neural closed-loop sensor simulator.
\newblock In \emph{Proceedings of the IEEE/CVF Conference on Computer Vision and Pattern Recognition}, pages 1389--1399, 2023{\natexlab{b}}.

\bibitem[Yang et~al.(2024)Yang, Gao, Zhou, Jiao, Zhang, and Jin]{yang2024deformable}
Ziyi Yang, Xinyu Gao, Wen Zhou, Shaohui Jiao, Yuqing Zhang, and Xiaogang Jin.
\newblock Deformable 3d gaussians for high-fidelity monocular dynamic scene reconstruction.
\newblock In \emph{Proceedings of the IEEE/CVF Conference on Computer Vision and Pattern Recognition}, pages 20331--20341, 2024.

\bibitem[Yu et~al.(2024{\natexlab{a}})Yu, Xing, Yuan, Hu, Li, Huang, Gao, Wong, Shan, and Tian]{yu2024viewcrafter}
Wangbo Yu, Jinbo Xing, Li Yuan, Wenbo Hu, Xiaoyu Li, Zhipeng Huang, Xiangjun Gao, Tien-Tsin Wong, Ying Shan, and Yonghong Tian.
\newblock Viewcrafter: Taming video diffusion models for high-fidelity novel view synthesis.
\newblock \emph{arXiv preprint arXiv:2409.02048}, 2024{\natexlab{a}}.

\bibitem[Yu et~al.(2023)Yu, Xu, Zhang, Liu, Ye, Wu, Yan, Zhu, Xiong, Liang, et~al.]{yu2023mvimgnet}
Xianggang Yu, Mutian Xu, Yidan Zhang, Haolin Liu, Chongjie Ye, Yushuang Wu, Zizheng Yan, Chenming Zhu, Zhangyang Xiong, Tianyou Liang, et~al.
\newblock Mvimgnet: A large-scale dataset of multi-view images.
\newblock In \emph{Proceedings of the IEEE/CVF conference on computer vision and pattern recognition}, pages 9150--9161, 2023.

\bibitem[Yu et~al.(2024{\natexlab{b}})Yu, Wang, Yang, Wang, Xie, Cai, Cao, Ji, and Sun]{yu2024sgd}
Zhongrui Yu, Haoran Wang, Jinze Yang, Hanzhang Wang, Zeke Xie, Yunfeng Cai, Jiale Cao, Zhong Ji, and Mingming Sun.
\newblock Sgd: Street view synthesis with gaussian splatting and diffusion prior.
\newblock \emph{arXiv preprint arXiv:2403.20079}, 2024{\natexlab{b}}.

\bibitem[Zhang et~al.(2023)Zhang, Rao, and Agrawala]{zhang2023adding}
Lvmin Zhang, Anyi Rao, and Maneesh Agrawala.
\newblock Adding conditional control to text-to-image diffusion models.
\newblock In \emph{Proceedings of the IEEE/CVF International Conference on Computer Vision}, pages 3836--3847, 2023.

\bibitem[Zhao et~al.(2024)Zhao, Ni, Wang, Zhu, Huang, Chen, Wang, Zhang, Mei, and Wang]{zhao2024drivedreamer4d}
Guosheng Zhao, Chaojun Ni, Xiaofeng Wang, Zheng Zhu, Guan Huang, Xinze Chen, Boyuan Wang, Youyi Zhang, Wenjun Mei, and Xingang Wang.
\newblock Drivedreamer4d: World models are effective data machines for 4d driving scene representation.
\newblock \emph{arXiv preprint arXiv:2410.13571}, 2024.

\bibitem[Zhou et~al.(2024{\natexlab{a}})Zhou, Shao, Xu, Bai, Qiu, Liu, Wang, Geiger, and Liao]{zhou2024hugs}
Hongyu Zhou, Jiahao Shao, Lu Xu, Dongfeng Bai, Weichao Qiu, Bingbing Liu, Yue Wang, Andreas Geiger, and Yiyi Liao.
\newblock Hugs: Holistic urban 3d scene understanding via gaussian splatting.
\newblock In \emph{Proceedings of the IEEE/CVF Conference on Computer Vision and Pattern Recognition}, pages 21336--21345, 2024{\natexlab{a}}.

\bibitem[Zhou et~al.(2018)Zhou, Tucker, Flynn, Fyffe, and Snavely]{zhou2018stereo}
Tinghui Zhou, Richard Tucker, John Flynn, Graham Fyffe, and Noah Snavely.
\newblock Stereo magnification: Learning view synthesis using multiplane images.
\newblock \emph{arXiv preprint arXiv:1805.09817}, 2018.

\bibitem[Zhou et~al.(2024{\natexlab{b}})Zhou, Lin, Shan, Wang, Sun, and Yang]{zhou2024drivinggaussian}
Xiaoyu Zhou, Zhiwei Lin, Xiaojun Shan, Yongtao Wang, Deqing Sun, and Ming-Hsuan Yang.
\newblock Drivinggaussian: Composite gaussian splatting for surrounding dynamic autonomous driving scenes.
\newblock In \emph{Proceedings of the IEEE/CVF Conference on Computer Vision and Pattern Recognition}, pages 21634--21643, 2024{\natexlab{b}}.

\bibitem[Zhou and Tulsiani(2023)]{zhou2023sparsefusion}
Zhizhuo Zhou and Shubham Tulsiani.
\newblock Sparsefusion: Distilling view-conditioned diffusion for 3d reconstruction.
\newblock In \emph{CVPR}, 2023.

\end{thebibliography}
}


\end{document}